\documentclass[10pt,conference]{IEEEtran}
\IEEEoverridecommandlockouts
\usepackage{caption}
\usepackage{cite}
\usepackage{amsmath,amssymb,amsfonts} 
\usepackage[numbers,sort&compress]{natbib}
\usepackage{ulem}
\usepackage{algorithmic}
\usepackage{graphicx}
\usepackage{textcomp}
\usepackage{xcolor}
\usepackage{booktabs}
\usepackage{tikz}
\usepackage{markdown}
\newcommand*{\circled}[1]{\lower.7ex\hbox{\tikz\draw (0pt, 0pt)%
    circle (.5em) node {\makebox[1em][c]{\small #1}};}}
\def\BibTeX{{\rm B\kern-.05em{\sc i\kern-.025em b}\kern-.08em
    T\kern-.1667em\lower.7ex\hbox{E}\kern-.125emX}}
\usepackage[ruled]{algorithm2e}
\usepackage{multirow}
\usepackage{threeparttable}
\begin{document}
\setlength{\intextsep}{5pt}

% \title{Untangling Commits By Detecting Hidden Dependencies With Fine-Grained, Global Context-Aware, HeterogeneoFor example,aph Model}

\title{A new approach for encoding code and assisting code understanding}

% \author{ANONYMOUS AUTHOR(S)}

\author{
\IEEEauthorblockN{Mengdan Fan}
\IEEEauthorblockA{\textit{Peking University} \\
Bejing, China \\
fanmengdan@pku.edu.cn}
\and
\IEEEauthorblockN{Wei Zhang\textsuperscript{*}\thanks{* Corresponding authors}}
\IEEEauthorblockA{\textit{Peking University} \\
Bejing, China \\
zhangw@sei.pku.edu.cn}
\and
\IEEEauthorblockN{Haiyan Zhao}
\IEEEauthorblockA{\textit{Peking University} \\
Bejing, China \\
zhhy.sei@pku.edu.cn}
\and
\IEEEauthorblockN{Zhi Jin\textsuperscript{*}}
\IEEEauthorblockA{\textit{Peking University} \\
Beijing, China \\
zhijin@pku.edu.cn}
}

\maketitle

\begin{abstract}
Some companies (e.g., Microsoft Research and Google DeepMind) have discovered some of the limitations of GPTs' autoregressive paradigm \emph{next-word prediction}, manifested in the model's lack of planning, working memory, backtracking, and reasoning skills. 
GPTs rely on a local and greedy process of generating the next word, without a global understanding of the task or the output.
We have confirmed the above limitations through specialised empirical studies of code comprehension. 
Although GPT-4 is good at producing fluent and coherent text, it cannot handle complex logic and generate new code that haven’t been seen, and it relies too much on the formatting of the prompt to generate the correct code.
We propose a new paradigm for code understanding that goes beyond the \emph{next-word prediction} paradigm, inspired by the successful application of diffusion techniques to image generation (Dalle2, Sora) and protein structure generation (AlphaFold3), which have no autoregressive constraints.
Instead of encoding the code in a form that mimics natural language, we encode the code as a \emph{heterogeneous image} paradigm with a memory of global information that mimics both images and protein structures.
We then refer to Sora's CLIP upstream text-to-image encoder model to design a text-to-code encoder model that can be applied to various downstream code understanding tasks.
The model learns the global understanding of code under the new paradigm \emph{heterogeneous image}, connects the encoding space of text and code, and encodes the input of text into the vector of code most similar to it.
Using self-supervised comparative learning on 456,360 text-code pairs, the model achieved a zero-shot prediction of new data. 
This work is the basis for future work on code generation using diffusion techniques under a new paradigm to avoid autoregressive limitations.
\end{abstract}

\begin{IEEEkeywords}
next-word prediction, CLIP, global information
\end{IEEEkeywords}

\section{Introduction}
% 一些研究发现 GPT系列大模型的 自回归地预测下一个单词的范式 存在局限性，例如。。。
Some studies have found that there are several limitations of the GPT's autoregressive next-word prediction paradigm.
Microsoft Research notes that GPT-4 lacks planning, working memory, ability to backtrack, and reasoning abilities \cite{DBLP:journals/corr/abs-2303-12712}.
For example, although the model has enough knowledge to answer the question, the architecture of the GPT-4 cannot give the whole correct answer at once, but needs to be guided step by step to give the correct answer.  
The autoregressive nature of GPT-4 forces it to solve problems sequentially.
Google Deepmind notes that GPTs are not applicable to most areas of mathematics due to the high cost of converting human proofs into a machine-verifiable autoregressive format \cite{DBLP:journals/nature/TrinhWLHL24}.
And permuting the premise order can cause a performance drop of over 30 $\%$ \cite{DBLP:journals/corr/abs-2402-08939}.
When producing full natural-language proofs on a set of geometry problems from the IMO competitions, GPT-4 has a success rate of only 0 $\%$, often making syntactic and semantic bugs throughout its output, and showing little understanding of the geometry knowledge and the problem statements themselves.
In summary, GPTs rely on a local and greedy process to generate the next word, without a global understanding of the task or the output \cite{DBLP:journals/corr/abs-2303-12712}.

% 我们进一步在代码理解领域做了实证研究，发现自回归范式的很多缺点。例如。。。
We further do empirical studies in the field of code understanding and confirm the limitations of the autoregressive paradigm.
For example, in terms of code understanding and generation, GPT-4 cannot understand complex logic with multi-step operations and generates incomplete code.
Moreover, GPT-4 relies excessively on the form of the prompt. Even if the input text has the same meaning, but just different syntax, the generated code will be much different and often contain bugs.
GPT-4 is also not good at matrix operation. 
It often introduces new bugs while fixing matrix operation's bugs.
But after giving the prompt step by step, GPT-4 can generate a correct answer, which shows that the model is actually trained with enough knowledge, but the autoregressive paradigm makes it not answer the question well.
All of these issues confirm the limitations of GPT-4 noted by Microsoft and Google. 
The limitations makes it overly dependent on the given prompt, unable to understand the meaning of the code globally, and also unable to generate complete and correct code including multi-step operation/matrix operation at one time.

% Diffusion在图像/生命分子式生成领域的展现了优越性,并且不会有上述限制
Diffusion technology has made significant progress in the generation of images \cite{dalle2, sora} and life molecules \cite{AlphaFold3} recently. It is a wonder that the model does not have limitations like that of GPTs. The model can learn the global information of the picture/life molecule, and generate the picture/life molecule at one time instead of step by step.
Diffusion is a class of latent variable models inspired by considerations from non-equilibrium thermodynamics, which was originally used to generate high-quality images \cite{diffusion}.
In particular, the CLIP+diffusion model (Dalle2) can create new pictures that have never been seen in the data set according to the input text, such as \emph{panda mad scientist mixing sparkling chemicals, artstation} \cite{dalle2, sora}.
The model seems to really understand the image.
Where CLIP is a transferable visual models from natural language supervision, which provides text embedding vector for diffusion model to generate high-quality images \cite{clip}.
The core idea of CLIP is to embed images and texts into a shared semantic space so that the embeddings of similar images and texts are closer in this space.

We expect that the CLIP+Diffusion technique can be applied to code generation to avoid the limitations of the autoregressive paradigm and find that code has similar properties with both natural language and image.
Both natural language and code use tokens to construct sentences or programs, and both need to follow certain grammatical rules to be correctly understood; existing techniques for code generation have drawn on deep learning techniques (e.g., GPTs) for natural language. 
There are a lot of ambiguity in natural language, such as polysemy, context dependence, and so on. 
Out-of-vocabulary (OOV) problems often occur when the model is dealing with natural language, i.e. the vocabulary encountered is not in its predefined vocabulary.
However, the code has almost no ambiguity and can avoid the problem of OOV by dividing the namespace. So we think the code does not need to use the autoregressive paradigm \emph{next-word prediction}.
In addition, we find similarities between codes and images: color images consist of components such as yellow, blue, black, etc., and each component (e.g., yellow) contains different entities (e.g., light yellow, dark yellow, etc.).
The code consists of components such as classes, methods, variables, etc., and each component (e.g., method) contains different entities (e.g., method 1, method 2, etc.).
We call the property that an object is composed of different types of components as heterogeneity.

We propose a transferable pre-training CLIP model for code understanding, which is the basis of generating high-quality code using diffusion technology and other code understanding tasks.
For code encoding, we first broke the autoregressive encoding paradigm of GPTs imitating natural language, and propose a \emph{single-channel, one-dimensional, heterogeneous image} encoding paradigm imitating images. 
Because according to the human programming mind, code is written in a sequence, so we encode the code as a single-channel, one-dimensional, heterogeneous image.
Built-in or common classes, methods, attributes, operators, and other components of the code are encoded as fixed numerical IDs.
The different entities of each component (for example, the built-in methods print() and len() in Python) are encoded as similar IDs, in order to imitate similar pixels in the image.
User defined components only have unique IDs in the same namespace, and different entities of each component are also encoded as similar IDs.
This encoding scheme avoids the OOV problem, so we do not need to embed the tokens, ensuring that the new paradigm imitate images, not natural languages.
For pretraining, we propose a code encoder consisting of one-dimensional convolution and local pooling, which is suitable for the above code encoding paradigm.
The encoder not only learns the preceding part of the code of the next token like GPTs, but iteratively learns the global information of code with the memory of global information.

% 实验
% We evaluate the effectiveness and efficiency of the proposed approach. 
% Experiments on the common C\# and Java datasets with 1,612 and 14k \emph{tangled commits}, and manually validated datasets(MVD) with 600 commits show that \emph{HD-GNN}
% achieves better untangling results (i.e., an average of 25$\%$ and 19$\%$ enhancement of effectiveness for C\# and Java, compared to existing approaches) and far superior to existing approaches on MVD, without sacrificing time efficiency. 
% Overall, our work makes the following contributions:
% \begin{itemize}
% \item 
% A fine-grained hierarchical graph model of commits.
% The graph model captures the global context of the code changes by considering the context both have or have no obvious (direct or indirect) code dependency with the code changes. 
% \item
% A novel \emph{HD-GNN} which can not only learn the embeddings of heterogeneous directed graphs, but also learn hidden dependencies among code changes by aggregating the embeddings of the nodes and edges associated with the code changes in both connected and disconnected entity-level subgraphs.
% It can be extended to many other graph learning applications.  
% \end{itemize}

\section{Motivation}
GPTs' autoregressive next-word prediction paradigm has some limitations, such as lack of planning, working memory, ability to backtrack, and reasoning abilities, and do not apply to most areas of mathematics \cite{DBLP:journals/corr/abs-2303-12712, DBLP:journals/nature/TrinhWLHL24}.
Researchers believe that GPTs lack a global and deep understanding of the task or the output because they
rely on a local and greedy process of generating the next word \cite{DBLP:journals/corr/abs-2303-12712}.
In addition, during the long-term use of GPT-4 in the field of code understanding, we further confirmed the limitations through following cases studies.

\subsection{Cases Studies}

\subsubsection{Lack of ability to handle complex logic}
\begin{figure}[htbp]
\centerline{\includegraphics[width=0.45\textwidth]{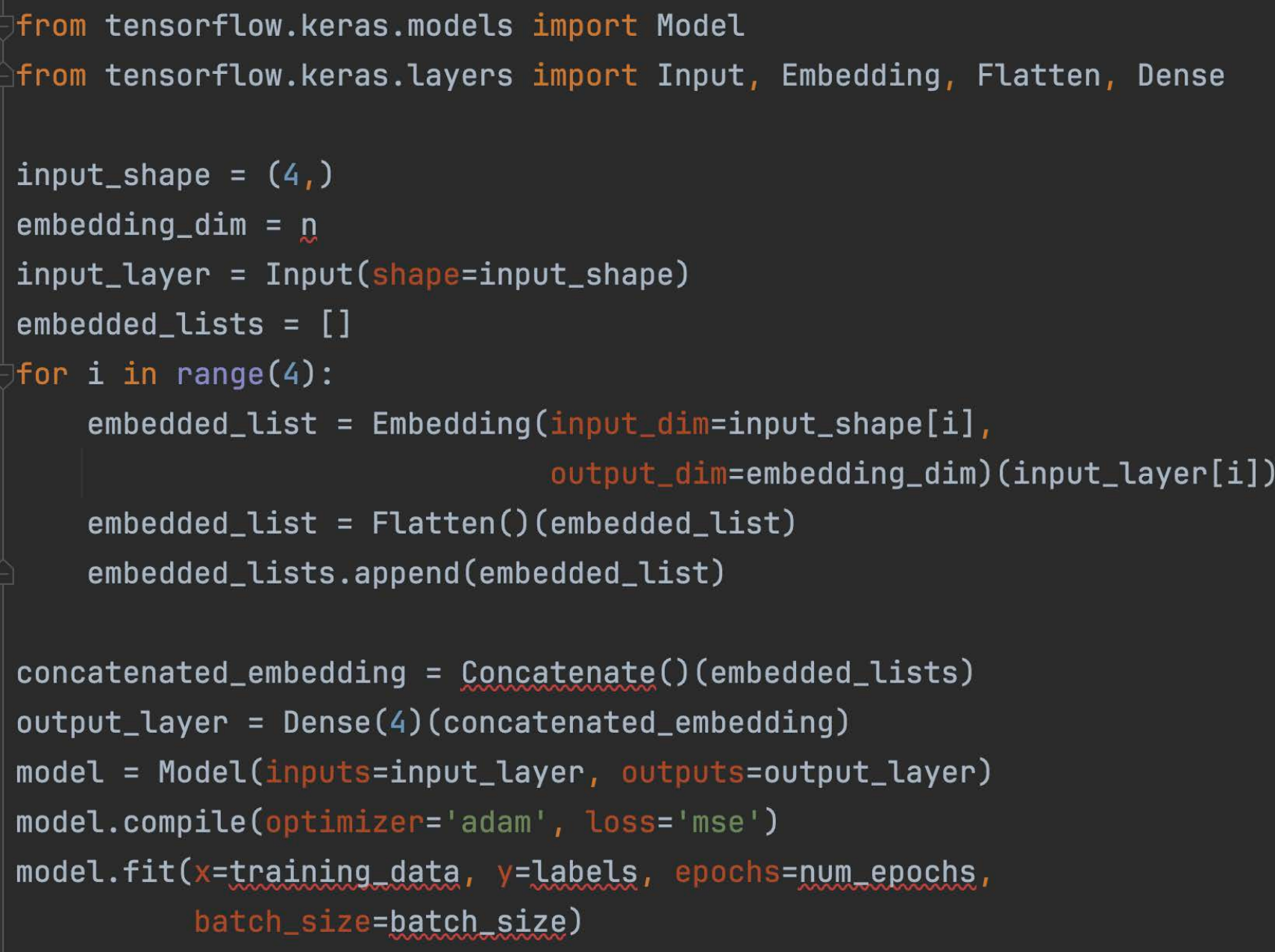}}
\caption{An example of GPT-4 handles complex logic.}
\label{complex_logic}
\end{figure}

We asked GPT-4 some code problems with complex logic and
multi-step operations, such as "My training data is ndarray with shape (359, 4), which means 359 arrays, and each array contains 4 lists (length is 1149, 3000, 3000, 18 respectively.).The label of the training data is ndarray with shape (359,).
Now, I want to train an embedding network whose input is the training data and whose output is an ndarray with shape (359, 4), representing 359 arrays, each containing 4 lists of length n. The output is an ndarray with shape (359, 4). How should it be implemented?
"
The statement may seem convoluted, so we have added a summarising "In a nutshell, we want the training data to be input into this embedding network, and each array contains 4 lists that are embedded to the same length n."
Although human beings can understand the meaning of the problem. GPT-4 gives us incomplete code full of logical bugs as shown in Fig. \ref{complex_logic}. E.g., GPT-4 does not understand the shape of the output we want (batch size, 4), and wrongly concatenates 4 lists of each item. The training process is also not complete, e.g., there is no \emph{for loop} of epochs. The variables indicated by red wavy line are also not explained. 

\subsubsection{Lack of working memory}
\begin{figure}[htbp]
\centerline{\includegraphics[width=0.45\textwidth]{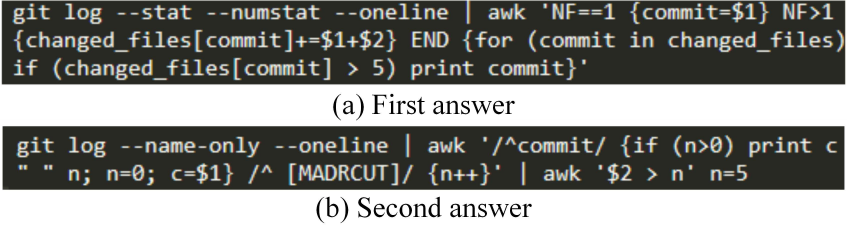}}
\caption{An example of GPT-4 lacking memory.}
\label{work_memory}
\end{figure}

If we ask the same question again after some rounds of dialogue because we forgot the early answer, GPT-4 often give a completely different answer. This will result in almost all dialogue rounds being invalid.
For example, we repeated the GPT-4 question "How do I use a terminal command to output a commit that changes more than n files in a repository?" twice, with several rounds of dialogue between them. GPT 4 answered two different wrong answers as shown in Fig. \ref{work_memory} (a) and (b) respectively.

\subsubsection{Overly dependent on prompt format}
\begin{figure}[htbp]
\centerline{\includegraphics[width=0.45\textwidth]{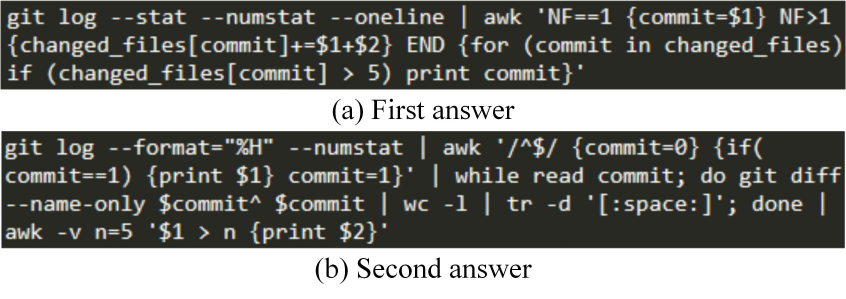}}
\caption{An example of GPT-4's over-reliance on the prompt format.}
\label{rely_prompt}
\end{figure}
GPT-4 will provide different answers to questions with different formats but the same meaning. For example, when we ask "How do I use a terminal command to find the commits that change more than n files in a repository?" and "How can I use a terminal command to find the commits with a number of changed files greater than n in a repository?" GPT-4 give very different answers as shown in Fig. \ref{rely_prompt}

\subsubsection{Not good at matrix operation}
\begin{figure}[htbp]
\centerline{\includegraphics[width=0.45\textwidth]{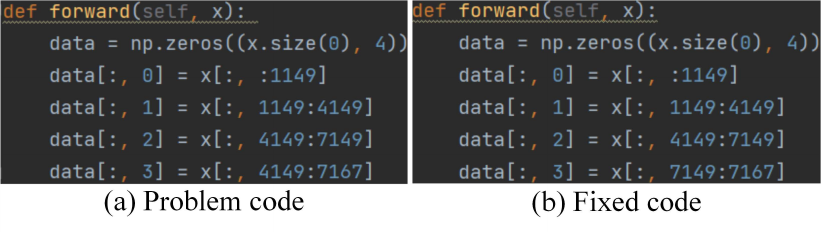}}
\caption{An example of GPT-4 is not good at matrix operation.}
\label{matrix}
\end{figure}

GPT-4 often fails to handle matrix operations correctly or introduces new bugs when handling matrix operations. For example, as shown in Fig. \ref{matrix}(a), when we asked GPT-4 why the figure (a) code reported an error: "ValueError: could not broadcast input array from shape (359,1149) into shape (359,)". GPT-4 will allow us to modify according to Fig. \ref{matrix}(b).
But it introduces new bugs into the code: \emph{ValueError: only one element tensors can be converted to Python scalars.}
GPT-4 ignores global correctness and instead gets stuck in solving local problems.

\subsubsection{Lack of creativity}
GPT-4 is not creative. For example, if we ask GPT-4 to invent an algorithm for predicting earthquakes, GPT-4 cannot provide an answer, even if it is to provide some ideas for this scientific question.

\subsection{Discussion}

\subsubsection{Using CLIP+Diffusion \cite{dalle2} in image understanding/generation can avoid some limitations similar to using GPTs in code understanding/generation}
\begin{figure}[htbp]
\centerline{\includegraphics[width=0.45\textwidth]{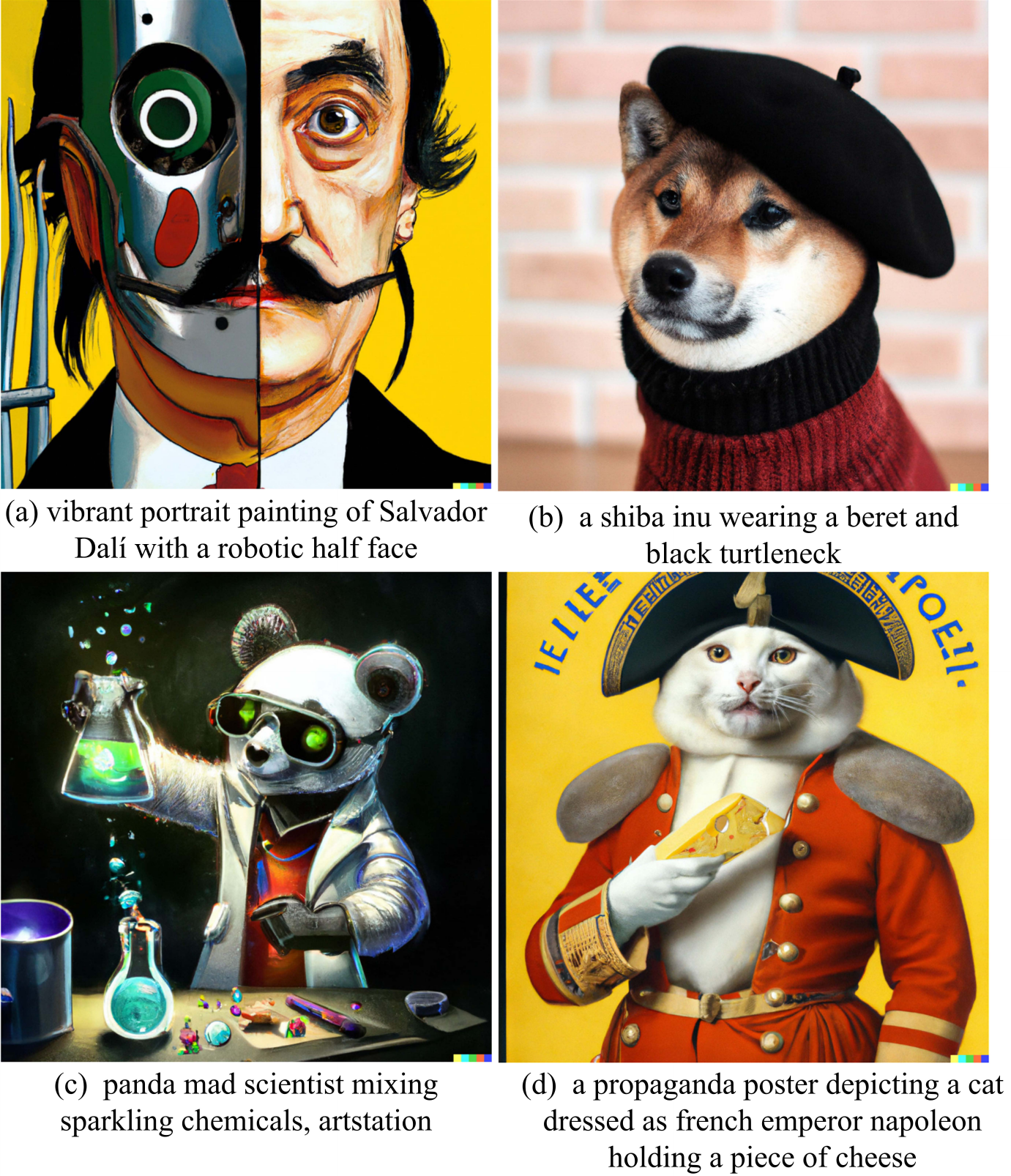}}
\caption{Examples of CLIP+Diffusion can avoid some limitations similar to GPTs.}
\label{discussion_p1}
\end{figure}

As shown Fig. \ref{discussion_p1}(a) and (b), even if a piece of text with complex logic is input,  CLIP+Diffusion can globally understand the text and correctly generate images with complex structures. 
The lines and colors in (a) are complex, precise, and smooth. The complex fur and texture in (b) are generated clearly.
Even if the form of the input text is changed, as long as the meaning is the same, the generated image is still correct.
In addition, CLIP+Diffusion also has creativity. 
As shown Fig. \ref{discussion_p1}(c) and (d), \emph{panda mad scientist} and \emph{a cat dressed as french emperor
napoleon holding a piece of cheese} hardly ever appear in real life and the training data. But CLIP+Diffusion technology can create unprecedented images based on human descriptions.
Therefore, we believe that if this technology is applied to code understanding/generation, it will greatly improve performance in this area.

\subsubsection{In contrast to CLIP+Diffusion for image understanding/generation, why do GPTs for code understanding/generation have the above limitations}
% Where CLIP is a transferable visual models from natural language supervision, which provides text embedding vector for diffusion model to generate high-quality images \cite{clip}.
% The core idea of CLIP is to embed images and texts into a shared semantic space so that the embeddings of similar images and texts are closer in this space.
We find that the CLIP model would memorize the entire image information globally while training, rather than only memorizing the previous code to infer the next token as GPTs do.
For example, CLIP learned about the global distribution of fur color and texture, as well as the structure and position information of facial features of many pictures of shiba inu. 
Thus, the image representation corresponding to the textual description given by the human output by CLIP already has global information about the image \cite{dalle2}. Then Diffusion further restores a high-quality image on this image representation.

However, due to GPTs adhering to the paradigm of predicting the next token, the model is unable to globally remember fully functional code information.
As shown in Fig. \ref{complex_logic}, if we ask GPT-4 to generate a logically complex code, even if GPT-4 has already learned the complex code, because it can only predict the next token based on the previous code, GPT-4 cannot generate a structurally complete code.
Other limitations of GPTs should also be caused by this paradigm.
Specifically, after a few rounds of dialogue, when we ask the same question again, due to changes in the previous text, the answers given by GPT-4 are completely different.
And, questions with the same meaning but different forms are different previous texts, so GPT-4 will generate different answers.
In addition, the reason why GPTs are not good at handling matrix operations and has creativity is because it does not truly understand the code. Understanding code requires learning the entire code information globally during the training process, rather than local context. Overall, we believe that GPTs' autoregressive \emph{next-word prediction} paradigm is the fundamental reason for the limitations of GPTs.

\subsubsection{Why is the autoregressive paradigm adopted by GPTs to understand/generate code}
Autoregressive paradigm is used for understanding and generating code by GPTs because it is mainly applicable to understanding and generating natural language. 
Both code and natural language are tools that people use to communicate. Natural language is used for communication between humans, while code is used for communication between humans and computers.
Both natural language and code are token sequences with specific syntax rules.
However, natural language is more flexible than code. Natural language can change according to different contexts, e.g. the meaning of words can change according to the context.
That is, the same words have ambiguity in different contexts.
Therefore, the understanding and generation of natural language require a strict reliance on contextual analysis. 
This is also the main reason why GPTs use autoregressive paradigm.
Although Transformer encoders can globally learn the information of the entire text and classify the predicted words, Transformer decoders still use the autoregressive paradigm in the generation process \cite{transformer}.
This paradigm generates tokens step by step to ensure semantic correctness, relying on the previous text for each step, rather than outputting complete sentences.
And GPTs only adopt the decoder of the transformer.
We summarize the above analysis in columns 1 and 3 of table \ref{similarity}.

\subsubsection{What are the challenges and solution in proposing a new paradigm that can avoid the limitations of GPTs}
To avoid the limitations of GPTs, we attempt to mimic images and propose a new structured coding paradigm for code, making it easier to apply CLIP+Diffusion technology.
But we found that the code not only has features of images but also features of natural language.
This has led to the challenge of proposing a new paradigm. 
As shown in columns 2 and 4 of table \ref{similarity}, in terms of imitating images, we find there are similarities between codes and images.
Images consist of components/colors (e.g., yellow, blue), and each component contains different entities (e.g., light yellow, dark yellow for yellow).
Code consists of components/types of entities (e.g., classes, methods, variables), and each component contains different entities (e.g., method 1, method 2 for method).
We call the property \emph{an object is composed of different types of components} is heterogeneity.

However, we found that different tokens in the code cannot be represented with finite numbers, leading to Out-of-vocabulary (OOV) issues.
OOV problem means that the vocabulary encountered is not in the predefined vocabulary, which is often the case when the model is dealing with natural language.
If the embedding technique in natural language is used to solve this problem, it will result in the inability of the CLIP+Diffusion technique to be applied. This is the challenge of proposing a new paradigm.
However, unlike natural language, the namespaces of entities in code could much smaller than those of words in natural language.
The namespace of entities in the code is related to the functional scope of the code, which can be packages, folders, files, or even code snippets.
However, the namespace of most words in natural language is often the entire vocabulary. 
Therefore, if the namespace is strictly partitioned and numbers are used repeatedly to encode these cross-namespace tokens, this may solve the OOV problem.

\begin{table}[]
\footnotesize
\caption{The similarity of image, text, and code}
\label{similarity}
\begin{center}
\begin{threeparttable}
\begin{tabular}{ccccc}
\hline
& Linguistic & Heterogeneous & Ambiguous & OOV \\ \hline
Image & $\times$ & $\checkmark$ & $\times$ & $\times$ \\
Code & $\checkmark$ & $\checkmark$ & $\times$ & Solvable \\
Text  & $\checkmark$ & $\times$ & $\checkmark$ & $\checkmark$ \\ \hline
\end{tabular}
\begin{tablenotes}
\footnotesize
\item
Linguistic: Is it language?
\item
Heterogeneous: Is it heterogeneous?
\item
Ambiguous: Is it possible to have ambiguity?
\item 
OOV: Is it has OOV issues?
\end{tablenotes}
\end{threeparttable}
\end{center}
\end{table}

\section{Approach}

\subsection{Heterogeneous image paradigm}
\begin{figure}[htbp]
\centerline{\includegraphics[width=0.4\textwidth]{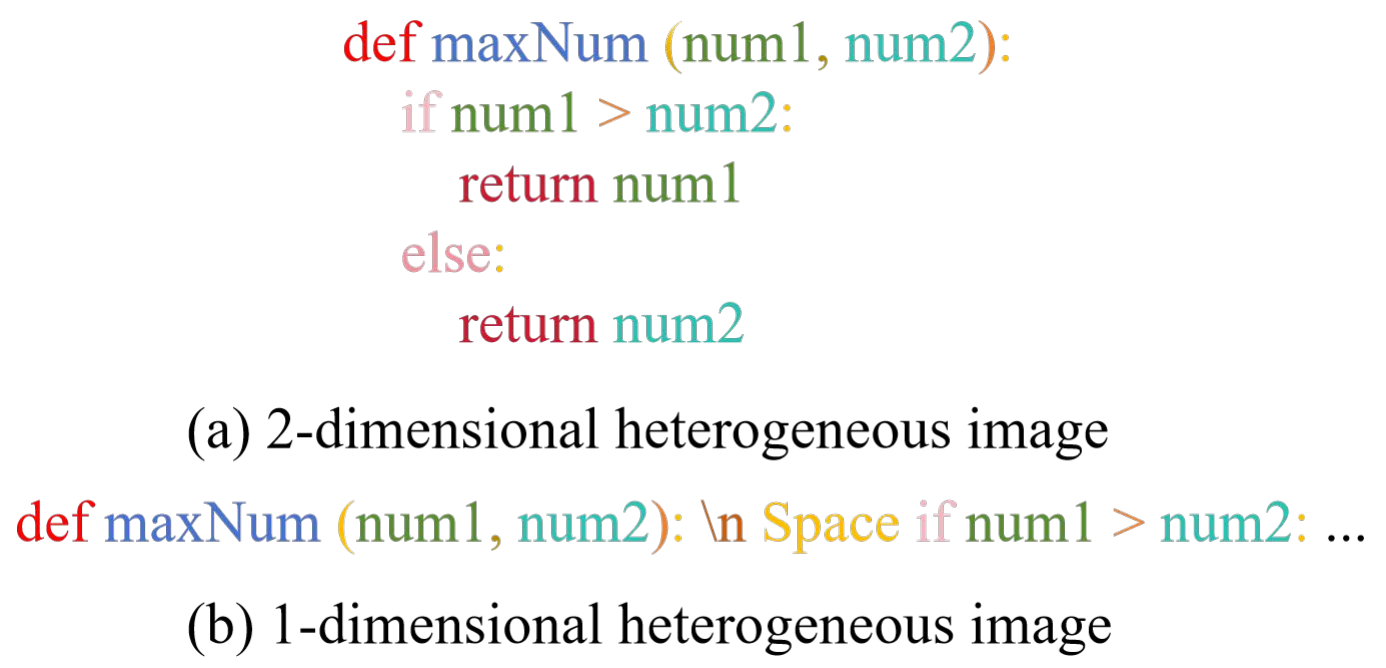}}
\caption{Examples of heterogeneous image.}
\label{heterogeneous_image}
\end{figure}

The code has a highly standardized structure and is not as flexible as the structure of natural language, so we first break the commonly used autoregressive \emph{next-word prediction} paradigm for code understanding and generation, and propose a novel single-channel, one-dimensional, \emph{heterogeneous image} paradigm.
Both code and image have heterogeneity, meaning that both image and code are composed of different components.
Therefore, we refer to images and consider different components in the code (e.g., classes, methods, variables, operators, numbers, symbols) as different components in the image (e.g., red, yellow, blue, green, etc.).
Different entities within the same component, e.g., class 1 and class 2, etc. in class) are represented by similar numerical values (IDs) to simulate the pixel values of different entities within the same component in the image (e.g., dark red, light red, pink, etc. in red).
To prevent significant differences in the values assigned to tokens due to too many different tokens (OOV problem), we take a series of measures:
1) Clean the code. Replace the messy data that will affect the accuracy of code tokenization with placeholders of different categories. For example, the messy string output by the print function is occupied with 'STR'. 
This step is optional and needs to be ensured that removing such messy data will not reduce the accuracy of the text-to-code matching results.
2) Divide tokens of entities such as the definition of classes, methods, and variables into built-in tokens in the code library and user-defined tokens,
3) Divide User-defined tokens into namespaces,
4) The built-in tokens are represented by fixed numerical values (IDs),
5) User-defined tokens are only represented as temporary values (can be reused within other namespaces) in the namespace.
As shown in Fig. \ref{heterogeneous_image}(a), a function is first represented as a 2-dimensional heterogeneous image, where keywords are represented by different similar red pixel values (e.g. bright red, pink, light red, etc.), variables are represented by similar green pixel values, and operators/symbols (e.g. \emph{\textgreater}, \emph{:}, \emph{(}) are represented by similar yellow pixel values.

In addition, we also follow the same process 1)-5) for tokens that represent entity calls (such as calls to methods or class attributes).
In order to further reduce the number of such tokens and avoid OOV issues, we abstracted these types of token. 
That is, we use the same numerical value to indicate the tokens that contain the same entity call, e.g., \emph{a\_param.strip} and \emph{address.strip} use the same numerical value to indicate.
We then maintain a list for each such numerical value for users to manually find which specific token it is.
Using the encoding method described above, we created a Python vocabulary on the 456,360 code snippets in the training set in the CodeSearchNet data set\footnote{https://opendatalab.org.cn/CodeSearchNet}.
CodeSearchNet corpus contains code snippets and associated document comments from various programming languages from open-source repositories on GitHub. This corpus has been initiated and maintained by GitHub to encourage the development of code search and code understanding technologies.
The number of tokens and the corresponding IDs are different in different programming language. 
For example, the specific components and their numerical coding ranges in Python are shown in the Tab. \ref{Component_ID}.

\begin{table}[]
\footnotesize
\caption{Components and their ID ranges}
\label{Component_ID}
\begin{center}
\begin{threeparttable}
\begin{tabular}{ll|ll}
\hline
Components               & ID Range     & Components               & ID Range       \\ \hline
Keyword             & 1-35      & Built-in Attribute & 6930-7960   \\
Built-in Class      & 36-54     & Variable           & 7961-9999   \\
Class               & 55-1584   & Built-in AttrCall  & 10000-11270 \\
Built-in Method     & 1585-2698 & Attribute Call     & 11271-11509 \\
Method              & 2699-4454 & Operator           & 11510-11554 \\
Built-in MethCall & 4455-6128 & Number             & 11555-13811 \\
Method Call         & 6129-6929 &                    &             \\ \hline
\end{tabular}
\begin{tablenotes}
\footnotesize
\item
AttrCall: Attribute call,
MethCall: Method call.
\end{tablenotes}
\end{threeparttable}
\end{center}
\end{table}

For color images, there are three primary colors (red, blue, green), and other colors can be composed of these three primary colors. So, a color image consists of three channels.
The heterogeneous image proposed in this paper has only a single channel. Different components are equally important, rather than the fact that most components/colors in the image are composed of three basic components/colors.
According to human programming thinking, code is written sequentially.
This means that the dependence of each token on the tokens above and below it is not as high as the dependence on the tokens before and after it. 
Thus, the two-dimensional heterogeneous image is finally flattened into a one-dimensional heterogeneous image, and line breaks and spaces are retained and treated as symbols, as shown in Fig. \ref{heterogeneous_image}(b).
This structure also looks like a living molecule chain (proteins, DNA, RNA, etc.). 
For example, they are all composed of basic elements and have heterogeneity, and they also look like a chain.
DeepMind has applied diffusion technology to predict the structure of living molecules. 
This further inspires us to explore and promote the application of diffusion technology in code understanding and generation. Designing a robust pre-trained model for text-to-code is the most fundamental task.
Overall, code has similarities with both images and natural language. 
The new paradigm simulates the numerical encoding of pixels in images on the token's numerical encoding in code. 
But in terms of shape, the new paradigm still looks like a sequence of natural language.

\subsection{Contrastive Language-Code Pre-training} 

\subsubsection{Architecture}
\begin{figure*}[htbp]
\centerline{\includegraphics[width=1\textwidth]{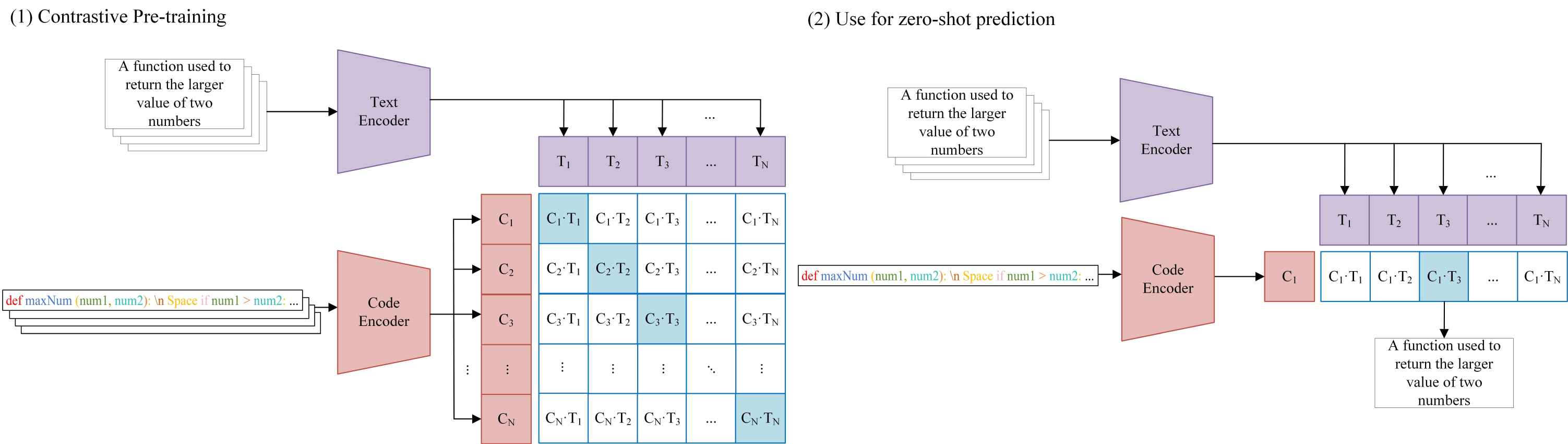}}
\caption{Architecture.  
CLCP jointly trains a code encoder and a text encoder to predict the correct(code, text) pairs. In the testing phase, for the target code, the learned model is treated as a zero shot classifier to select the description that best matches it.}
\label{Architecture}
\end{figure*}

We adopt a similar architecture of the Contrastive Language Image Pre-training (CLIP) model \cite{clip}, and propose the Contrastive Language Code Pre-training (CLCP) model. 
During the training phase, we jointly trains an code encoder and a text encoder to predict the correct (code, text) pairs of a batch.
As shown in Fig. \ref{Architecture},
given a batch of $N$ (code, text) pairs, CLCP is trained to predict which of the $N\times N$ possible (code, text) pairings in the batch are the correct pairs.
CLCP learns a multi-modal embedding space by jointly training an code encoder and a text encoder.
In view of the strong advantages of the pre-trained transformer model in understanding natural language \cite{transformer, Bert, GPT3, Chatgpt}, we continue to train directly on the model with our data. However, because we have adopted a new code coding paradigm, we have designed a new code encoder, which will be introduced later.

The model parameters are updated by the loss function, which maximizes the cosine similarity $C_i\cdot T_i(1\le i\le N)$ of the code embeddings $C_i$ and the text embeddings $T_i$ of the correct pairs $N$ in the batch while minimizing the cosine similarity $C_i\cdot T_j(1\le i, j\le N, i\neq j)$ of the code embeddings $C_i$ and text embeddings $T_j$ of the incorrect pairings $N^2-N$.
Specifically, we optimize a symmetric cross entropy loss over these similarity scores \cite{clip}.
We expect these two encoders to be used for many downstream tasks for code understanding in the future, such as using a text encoder to encode input text into a vector in the multi-modal embedding space and then inputting that vector into the diffusion model to restore high-quality code.
Alternatively, use code encoders in different languages to translate code written in one programming language into code written in another programming language, etc.

\subsubsection{Code encoder based on one-dimensional convolution}

\begin{figure}[htbp]
\centerline{\includegraphics[width=0.5\textwidth]{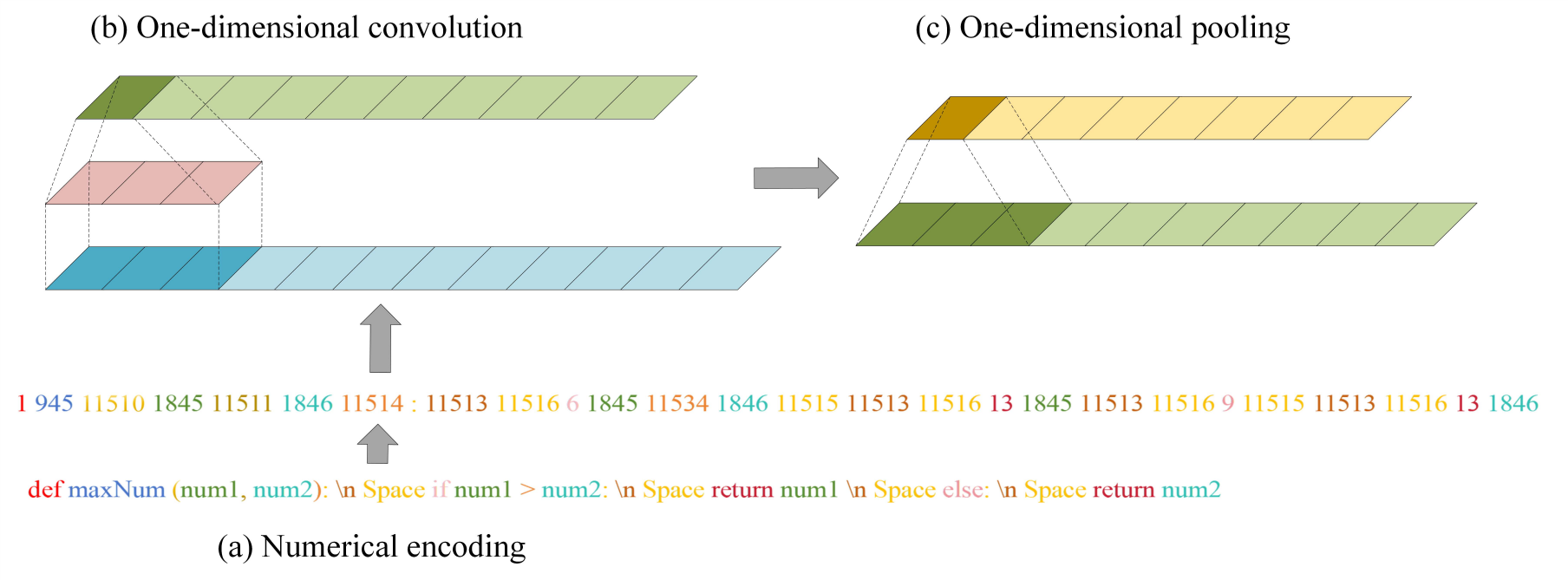}}
\caption{The specific implementation of a block of the code encoder.}
\label{Code_encoder}
\end{figure}

For the code encoder, we refer to two-dimensional convolution for learning on images \cite{CNN}, and design a one-dimensional convolutional and pooling neural network suitable for one-dimensional heterogeneous images. 
As shown in Fig. \ref{Code_encoder}, the source code is first encoded in one-dimensional heterogeneous images according to the new encoding paradigm.
Secondly, we use one-dimensional convolution to convolve one-dimensional heterogeneous images with a kernel size of $k$ and a step size of $s$ to learn the semantic and structural information of the code.
That is, the convolution kernel slides on the input one-dimensional image with a step size of $s$, and at each step the IDs in the one-dimensional image are multiplied by the corresponding weights in the convolution kernel and summed to obtain the IDs of the dimensions of the output new vector/feature map.
Each value of the new vector will be activated by the Relu function.
Third, a one-dimensional (1D) max or average pooling is applied to the activated new vector, with window size $k'$ and step size $s'$.
The basic idea of pooling operation is to divide the input 1D image (also known as a vector) into several sub-vectors and calculate the maximum or average value of each sub-vector as the value of each dimension of the output vector.
Overall, convolution operations are mainly used to learn semantic and structural information between code tokens, while pooling is mainly used to aggregate information from sub-code fragments.
Finally, the code encoder consists of $M$ blocks, each containing a convolution layer and a pooling layer.

\begin{figure*}[htbp]
\centerline{\includegraphics[width=0.8\textwidth]{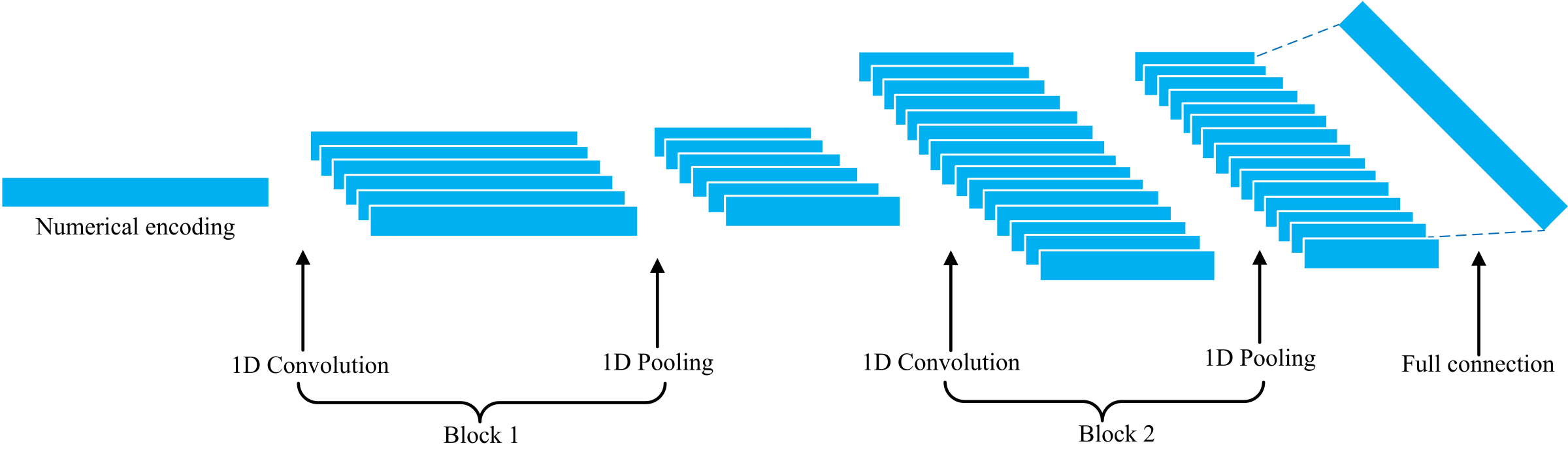}}
\caption{Overview of the code encoder.}
\label{Code_encoder_ov}
\end{figure*}

Fig. \ref{Code_encoder_ov} shows the overview of a code encoder containing two blocks.
Each convolution layer outputs
multiple feature maps, each of which is calculated by sliding a 1D convolution kernel on the input numerical encoding (i.e., the process in Fig. \ref{Code_encoder}(b)).
Each pooling layer receives the output of the convolution layer activated by Relu, performs pooling operations on them respectively, outputs the same number of feature maps, and then passes them to the next block.
The multiple feature maps output by the last block are flattened and input into a fully connected layer.

\subsubsection{Strategies for optimizing the training process}
We use Relu function \cite{Relu} to activate the output of the convolutional layer and use the He initialization \cite{kaiming} method to initialize the parameters of the convolutional layer.
Compared with traditional Sigmoid and Tanh activation functions, Relu has a linear relationship when the input is greater than 0, which means it will not saturate (i.e. the gradient will not approach 0). This helps to solve the problem of gradient vanishing, especially in deep networks \cite{af_survey,af_survey2,GELU}.
However, in deep networks, if the weight values are large, the accumulation of gradients during backpropagation can lead to gradient explosion problems \cite{GELU,kaiming}. 
So we use He Initialization \cite{kaiming} to avoid the above problem.
The core idea of Kaiming initialisation is to adjust the initial values of the weights based on the number of input nodes, in order to maintain consistency in the variance of the activation values across all layers.
Specifically, if a neuron has $n$ input nodes, its weights are initialised to a Gaussian or uniform distribution with a mean of $0$ and a variance of $2/n$.
The advantage of this is that no matter how deep the network is, the distribution of activation values will remain within a reasonable range, avoiding the problem of vanishing or exploding gradients and making the network easier to train.

In addition, Batch Normalization (BN) is a commonly used technique that normalizes each batch of data.
The BN layer can reduce the data distribution changes caused by parameter variations in the network, thereby avoiding gradient vanishing or exploding problems during training and improving the convergence performance of the model.
However, by normalizing the input of each layer, BN may suppress the diversity of some representations \cite{BN_dis} because BN forces each layer's input to have similar mean and variance during training, which may result in the model losing sensitivity to some important changes and diversity in the input distribution. 
This means that BN may not be applicable to certain types of neural networks, especially those that require preserving certain characteristics of the input distribution.
For example, CNN models used for image classification typically have the first few layers responsible for extracting low-level features of the image, such as edges, corners, textures, etc. In these layers, each feature map may focus on a specific pattern within the image.
However, BN independently normalizes each batch of feature maps, which may disrupt the relative differences between feature maps.
Therefore, when learning the code Heterogeneous image in this paper, we do not use BN to avoid blurring the distribution characteristics of the input image.
But in RQ2 of section \uppercase\expandafter{\romannumeral4} below, we will compare the effects of using BN and not using BN.

\section{Experiments}

\subsection{Research Questions}
% 不同大小的 训练集和测试集
% 不同的 code encoder变体
% 不同的 prompt engineering
The goal of this paper is to propose a new code encoding paradigm and design an applicable text-code encoder training model that can be migrated to different downstream tasks of code understanding so that the model may avoid the limitations of the autoregressive model on code understanding.
Following questions are set on 4$\times$NVIDIA Tesla V100 with 32GB RAM:
\textbf{RQ1. How effective is CLCP on Zero-Shot Transfer?}
It is difficult for the proposed CLCP to achieve high accuracy on the zero-shot task with a limited amount of data (456,360 (code, text) pairs) compared to the dataset (400 million (image, text) pairs) used for the CLIP model. 
For example, some previous work also used natural language supervision for image representation learning like CLIP \cite{DBLP:conf/iccv/LiJJM17, VirTex, ICMLM, DBLP:conf/mlhc/0004JMML22}.
However, due to the limited amount of training data, their effects are not as amazing as those of the CLIP (e.g., Li et al \cite{DBLP:conf/iccv/LiJJM17} reach only $11.5\%$ accuracy on ImageNet in a zero-shot setting), so they have not attracted enough attention.
Collecting a large amount of high-quality data is still an arduous task in the future.
In this paper, to evaluate whether the proposed method has application value, we change the size of training and testing sets to magnify the effect of the model on the zero-shot task and evaluate whether the model can effectively learn the global information of the code.

% \textbf{RQ2. How effective are different variants of CLCP?}
% We also evaluate the effect of different variants of CLCP to explore how to design the code encoder more reasonably.
\textbf{RQ2. Ablation experiment.}
In this RQ, we conducted some ablation experiments to explore training strategies for learning code heterogeneous images, rather than training strategies that are applicable to all models.
Specifically, pooling operations reduce the dimensionality of feature maps by aggregating information from adjacent regions, which inevitably results in the loss of some detailed information. For example, in max/average pooling, only the max/average value within each region is retained, while all other values are discarded.
This type of information loss can gradually accumulate in multi-layer networks, leading to poor model performance when fine features are required.
Therefore, in this RQ, we removed the pooling layer of CLCP model to verify its effectiveness.

In addition, we use He initialization \cite{kaiming} to avoid gradient explosion problems in backpropagation, and do not use BN to avoid blurring the distribution characteristics of the input image.
In this RQ, we remove He initialization and add BN respectively to verify the effectiveness of the optimization strategy we adopted during the training process.

\textbf{RQ3. Will the prompt engineering improve the results?}
Prompt engineering is a common method to improve the effect of model based on natural language supervision \cite{transformer, Bert, GPT3, clip}.
We observed that the description of the code in the dataset is complex and redundant.
The common categories of redundancy is summarized in table \ref{redundant_infor}.
So we try to clean up and reconstruct the text to improve the experimental results.

\subsection{Dataset}
Different program languages need different code parsers to make vocabularies. 
Therefore, the number of tokens and the corresponding IDs are different in different programming languages. 
To reduce the workload of program language preprocessing, this paper makes an exemplary evaluation in Python. 
We use the Python data set of the CodeSearchNet corpus\footnote{https://opendatalab.org.cn/CodeSearchNet}, including 456,360 (code, text) pairs in the training set and 22176 pairs in the testing set.
The CodeSearchNet corpus contains multiple public code libraries from GitHub, covering a variety of programming languages (such as Java, python, JavaScript, etc.).
The public code libraries contain a large number of code snippets from a variety of open-source software repositories, and the document comments corresponding to the code snippets help to understand the function and purpose of the code. 
This corpus has been initiated and maintained by GitHub to encourage the development of code-understanding technologies.
We selected 13,760 samples in the testing set that were completely different from the training set category by identifying different text descriptions and manually verifying them. These samples were used as an initial data set to evaluate the accuracy of the model in zero-shot tasks.

\subsection{Experimental Setup}

Because the amount of data is limited, we imitate the works \cite{DBLP:conf/iccv/LiJJM17, VirTex, ICMLM, ConVIRT} referred to in the clip paper to do some preliminary exploration to verify the proposed approach is effective (e.g. the accuracy in \cite{DBLP:conf/iccv/LiJJM17} is only 11.5$\%$), instead of pursuing the effect of training under 400 million data as in the clip model.

\subsubsection{Settings for RQ1}

\textbf{Dataset size:}
We observe whether the proposed approach is effective on the zero-shot transfer task by gradually increasing the size of the dataset and the depth of the model.
As for data set size, each time we randomly sample datasets of different sizes to evaluate the improvement of the model in zero-shot transfer compared to the results of random prediction. 
Specifically, the size of the training set is 30000, 60000, 120000, 240000, and 456,360 respectively. 456,360 is the size of the entire training set in the dataset.
For the above training set, we conducted two experiments. The testing set for the first experiment consists of 50 randomly selected samples. 
The testing set for the second experiment consists of 50, 100, 300, 600 and 1000 randomly selected samples. 

\textbf{Baselines:} 
\emph{CLCP$_{lp}$}: CLCP using local pooling layers. 
For the different sizes of training/testing sets, this type of CLCP is divided into five subtypes: CLCP$_{lp}$(3), CLCP$_{lp}$(4), CLCP$_{lp}$(5),  CLCP$_{lp}$(6),  CLCP$_{lp}$(7) which represent CLCP$_{lp}$ with 3, 4, 5, 6 and 7 blocks. 
A block indicates a functional unit that includes a convolution layer and a pooling layer.
\emph{CLCP$_{gp}$}: Similarly, CLCP uses global pooling layers. This type of CLCP is divided into five subtypes: CLCP$_{gp}$(3), CLCP$_{gp}$(4), CLCP$_{gp}$(5), CLCP$_{gp}$(6), and CLCP$_{gp}$(7).
% \emph{CLCP-WP}: CLCP without pooling layer. 
% Similarly, this type of CLCP is divided into five subtypes: CLCP-WP(3), CLCP-WP(4), CLCP-WP(5), CLCP-WP(6) and CLCP-WP(7).
\emph{CLCP$_{rn}$}: 
Because ResNet \cite{ResNet} has achieved advanced performance in image coding and has been applied to the CLIP model, we use the same architecture as ResNet but replace the 2D convolution with 1D convolution. 
That is, the model consists of an input layer, $N$ residual blocks and a global pooling layer. Each block contains three convolution layers. 
Among them, the first two convolution layers process the input data in series, while the third convolution layer is used to convolute the input by $1\times1$ convolution directly. 
Similarly, for different sizes of training/testing sets, $N$ is taken as 3, 4, 5, 6 and 7.
Finally, the output of the third convolution layer is added to the output of the second convolution layer as the final output of a residual block.

\textbf{Evaluation Metrics:}
\emph{Accuracy (Acc.):} The proportion of the correctly matched (code, text) pairs in the $L$ testing pairs. 
If all L codes are randomly matched with the texts, the estimated acc (EA) is $1/L$.

% The total probability of all codes successfully matching their texts/descriptions is $(1/L)^L$.
% Given the code in the test data, the probability of successfully matching the pairs of $l$ (code, text) after randomly selecting a text for each code is $(P = \binom{L}{l} \times (1/L)^l \times ((L-1)/L)^{(L-l)})$.

% \subsubsection{Settings for RQ2}
% \textbf{Baselines:} 
% \emph{CLCP$_{lp}$}: CLCP using local pooling layers. Similarly, this type of CLCP is divided into five subtypes: CLCP$_{lp}$(3), CLCP$_{lp}$(5), and CLCP$_{lp}$(7).
% \emph{CLCP$_{gp}$}: CLCP using global pooling layers. This type of CLCP is divided into five subtypes: CLCP$_{gp}$(3), CLCP$_{gp}$(5) and CLCP$_{gp}$ (7), which represent CLCP$_{gp}$ with 3, 5 and 7 layers of convolution layers.
% \emph{CLCP-WP}: CLCP without pooling layer. Similarly, this type of CLCP is divided into five subtypes: CLCP-WP(3), CLCP-WP(5) and CLCP-WP(7).
% \textbf{Evaluation Metrics:}
% Use the same metrics as those in RQ1 here.

%%%%%%%%%%%%%%%%%%%%%%%%%%%%%%%%%%%%%%%%%%%%%%
\subsubsection{Settings for RQ2}

\textbf{Dataset size:}
We use the same dataset sizes in RQ1.

\textbf{Baselines:} 
Two most effective baselines in RQ1 are used, CLCP$_{lp}$ and CLCP$_{rn}$ with different blocks.
In addition, their variants are also used as baselines:
\emph{CLCP$_{lp}$+BN}: A variant of CLCP$_{lp}$ where the input of each convolutional layer is normalised by BN.
\emph{CLCP$_{lp}$-Pool}: A variant of CLCP$_{lp}$ where the pooling layer is removed.
\emph{CLCP$_{lp}$-Init}: A variant of CLCP$_{lp}$ where the He initialization is removed.
\emph{CLCP$_{rn}$+BN}: A variant of CLCP$_{rn}$ where the input of each convolutional layer is normalised by BN.
\emph{CLCP$_{rn}$-Pool}: A variant of CLCP$_{rn}$ where the pooling layer is removed.
\emph{CLCP$_{rn}$-Init}: A variant of CLCP$_{rn}$ where the He initialization is removed.

\textbf{Evaluation Metrics:}
In this RQ, we calculate the difference between the average accuracy of CLCP$_{lp}$/CLCP$_{rn}$ with different blocks and the average accuracy of their variants across all sizes of datasets to determine the variation in the effect of the variants relative to the original model.

%%%%%%%%%%%%%%%%%%%%%%%%%%%%%%%%%%%%%%%%%%%%%%
\subsubsection{Settings for RQ3}

\textbf{Data quality:}
The text that describes the code in CodeSearchNet contains a lot of redundant information. 
For example, it contains user-defined paths, folder structures, and some demonstration information containing code results with special symbols. 
The redundant information can reduce the effectiveness of the model. 
Therefore, we perform prompt engineering on the dataset, i.e. we filter the redundant information in the text. The specific categories of redundant information are listed in Table \ref{redundant_infor}.
We call the dataset filtered by prompt engineering is called \emph{CodeSearchNet(p)}.
Both \emph{CodeSearchNet(p)} and \emph{CodeSearchNet} are used in RQ2.

\begin{table*}[]
\centering
\scriptsize
\caption{Categories of redundant information of the description of the code}
\label{redundant_infor}
\begin{tabular}{|l|l|l|}
\hline
Type & Examples & Explain 
\\ \hline
\multirow{3}{*}{Contains special symbols} & \begin{tabular}[c]{@{}l@{}}The algorithm used is based on streamlib's implementation of\\ "HyperLogLog in Practice: Algorithmic Engineering of a State\\ of the Art Cardinality Estimation Algorithm", available here\\ \textless{}https://doi.org/10.1145/2452376.2452456\textgreater{}\end{tabular} 
& \begin{tabular}[c]{@{}l@{}}The text contains URLs with special symbols.\end{tabular}
\\ \cline{2-3} 
& \begin{tabular}[c]{@{}l@{}}Return the union of this RDD and another one.\\ \&amp;gt;\&amp;gt;\&amp;gt; rdd = sc.parallelize({[}1, 1, 2, 3{]})\\ \&amp;gt;\&amp;gt;\&amp;gt; rdd.union(rdd).collect()\\ {[}1, 1, 2, 3, 1, 1, 2, 3{]}\end{tabular} 
& \begin{tabular}[c]{@{}l@{}}The text contains a running demonstration with special \\ symbols.\end{tabular}
\\ \cline{2-3} 
& \begin{tabular}[c]{@{}l@{}}Structure:\\train\_dir. person1 somename1.jpeg... somename1.jpeg ...\end{tabular}
& \begin{tabular}[c]{@{}l@{}}The text contains the file directory\end{tabular}
\\ \hline
\multirow{2}{*}{Unclear explanation}      
& \begin{tabular}[c]{@{}l@{}}{[}True, True, False, True, False{]}, 2 -\&gt;\\ {[}{[}True,  True{]}, {[}True, False{]}, {[}False, True{]}, {[}True, False{]},{]}\end{tabular} 
& \begin{tabular}[c]{@{}l@{}}The text is not clear, and only the running demonstration is \\ included.\end{tabular} \\ \cline{2-3} 
& \begin{tabular}[c]{@{}l@{}}Parameters\\ ----------\\ i : int, slice, or sequence of integers\\axis : int\end{tabular}  & \begin{tabular}[c]{@{}l@{}}The text is not clear. It all consists of phrases.\end{tabular}
\\ \hline
\end{tabular}
\end{table*}

\textbf{Dataset size:}
We use the same dataset sizes in RQ1 for \emph{CodeSearchNet} and \emph{CodeSearchNet(p)} respectively. 

\textbf{Baselines:} 
We use the two most effective baselines in RQ1, which are CLCP$_{lp}$ and CLCP$_{rn}$.
Similarly, for different sizes of training/testing sets of \emph{CodeSearchNet} and \emph{CodeSearchNet(p)}, CLCP$_{lp}$ is divided into CLCP$_{lp}$ and CLCP$_{lp}$(p), each of which is further divided into 5 subtypes that represent CLCP$_{lp}$ with 3, 4, 5, 6 and 7 blocks. 

\textbf{Evaluation Metrics:}
Use the same metrics of RQ1.

\subsection{Experiment Results}

\begin{figure}[!htbp]
\centerline{
\includegraphics
[width=0.35\textwidth]
{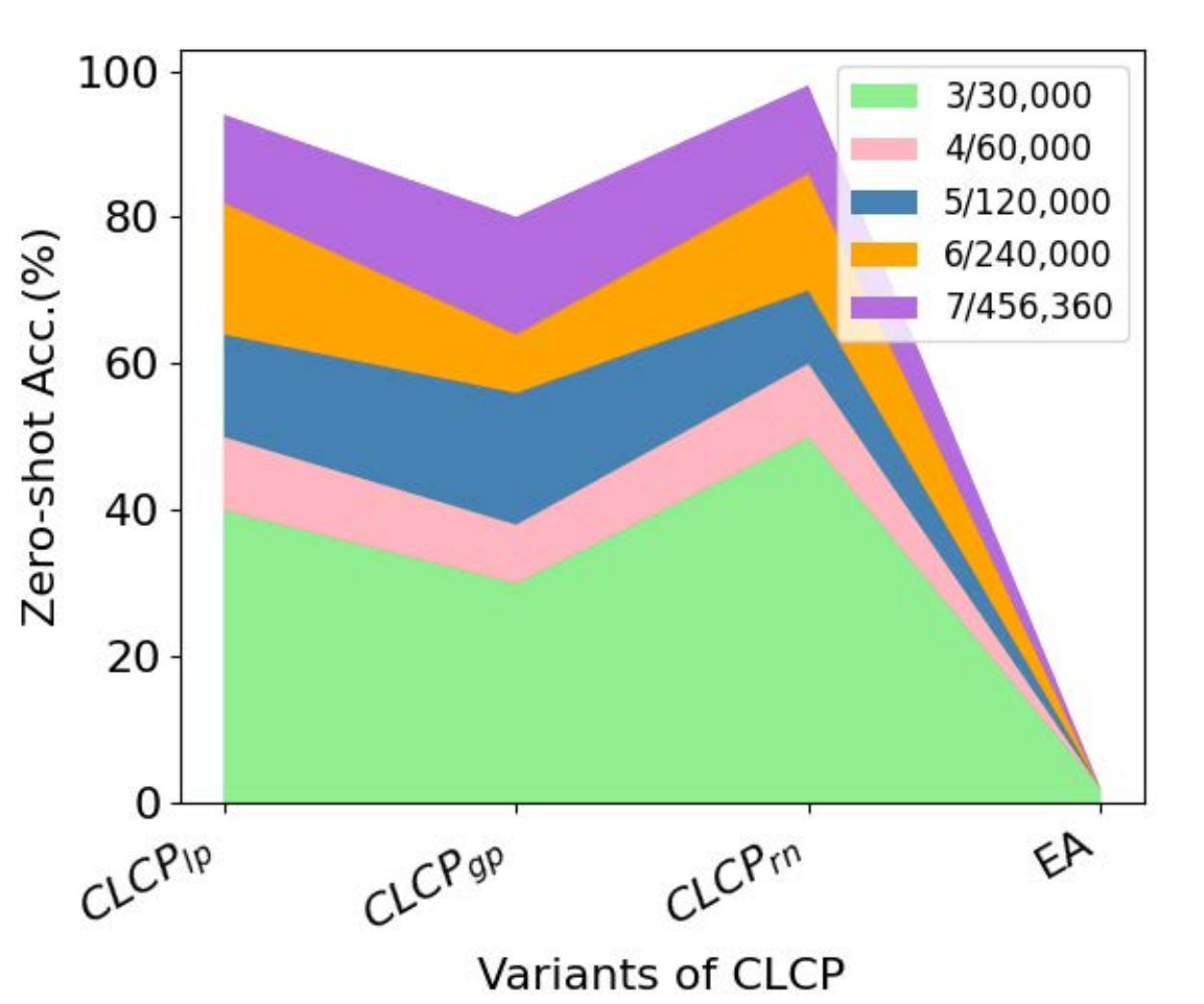}}
\caption{As the sizes of the training set and model increase, the performance of variants on the testing sets of same size.}
\label{RQ1_sameTest}
\end{figure}

\textbf{Answer to RQ1:}
As shown in Fig. \ref{RQ1_sameTest} (stacked area chart), the legend represents the size of the model/the size of the training set. The different colors represent the growth in the model accuracy in different model sizes and dataset sizes.
We can see that as the sizes of the training set and model increase, the performance of all models improves on the testing sets of different size.
And all models perform better than the estimated Acc(EA) of random prediction.
In particular, the overall effect of variants using global pooling is not as good as that of variants using local pooling.
This may be because global pooling compresses the entire feature map into a single value by averaging or maximizing operations, which to some extent loses some important local information. If this information is critical to the task, the effectiveness of global pooling will be reduced.
Local pooling is a pooling operation performed on a local region (usually a small sliding window) of a feature map, taking the maximum or average value within that region as the output, in order to preserve the salient information in the feature map.
Compared to global pooling, local pooling can retain more local information, which is more advantageous for learning code images that require fine feature representation.

\begin{figure}[!htbp]
\centerline{
\includegraphics
[width=0.35\textwidth]
{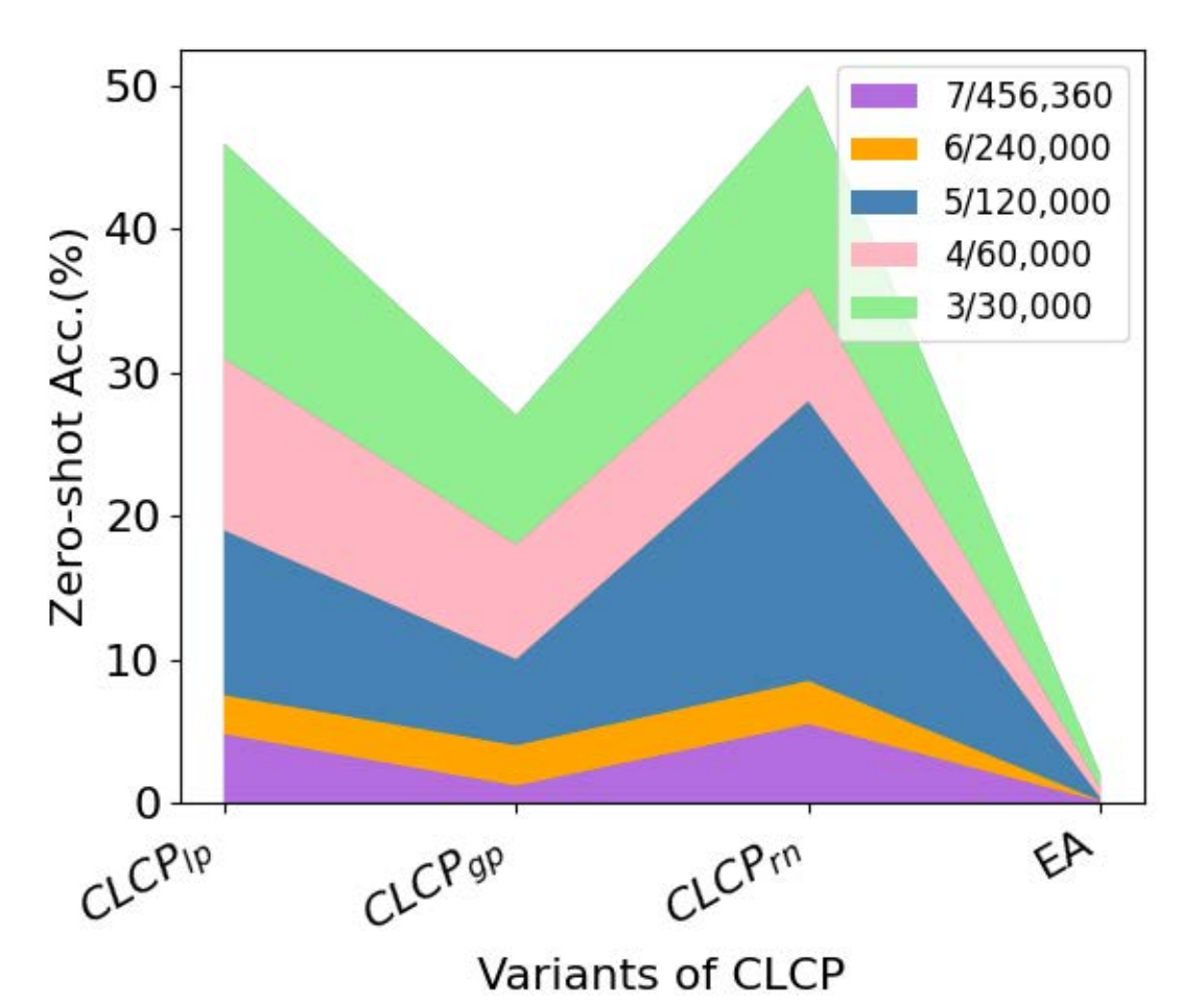}}
\caption{As the sizes of the training set and model increase, the performance of variants on the testing sets of different size.}
\label{RQ1_diffTest}
\end{figure}

For the results in Fig \ref{RQ1_diffTest}, when the size of the test set increases synchronously with the size of the training set and model, the performance of all variants of the model decreases.
That is to say, the model is underfitting.
This may be because the training data are too limited compared to CLIP, which was trained on 400 million data points. 
In particular, because we have not used embedding techniques in NLP but have redefined the paradigm of code representation, the model needs to learn the dependency relationships between tokens on its own. 
Therefore, this type of code image data has a high degree of diversity and complexity. 
So, the model may not be able to learn the general patterns of the data from a limited data.

% \begin{table}[htbp]
% \scriptsize
% \caption{Zero-shot performance on datasets of different size ($Accuracy\%$)}
% \label{RQ1_Acc}
% \begin{center}
% \begin{threeparttable}
% \begin{tabular}{c|ccccl}
% \hline
% \multicolumn{1}{l|}{\#Blocks} & \multicolumn{1}{l}{CLCP-AP} & \multicolumn{1}{l}{CLCP$_{gp}$} & \multicolumn{1}{l}{CLCP-WP} & \multicolumn{1}{l}{CLCP$_{rn}$} & EA                      \\ \hline
% 3                             & 40                          & 30                          & 30                          & 50                              & 10                      \\
% 4                             & 19                          & 10                          & 9                           & 28                              & 1                       \\
% 5                             & 4.8                         & 1.2                         & 1.1                         & 5.5                             & \multicolumn{1}{c}{0.1} \\ \hline
% \end{tabular}
% \end{threeparttable}
% \end{center}
% \end{table}

\begin{figure}[!htbp]
\centerline{
\includegraphics
[width=0.5\textwidth]
{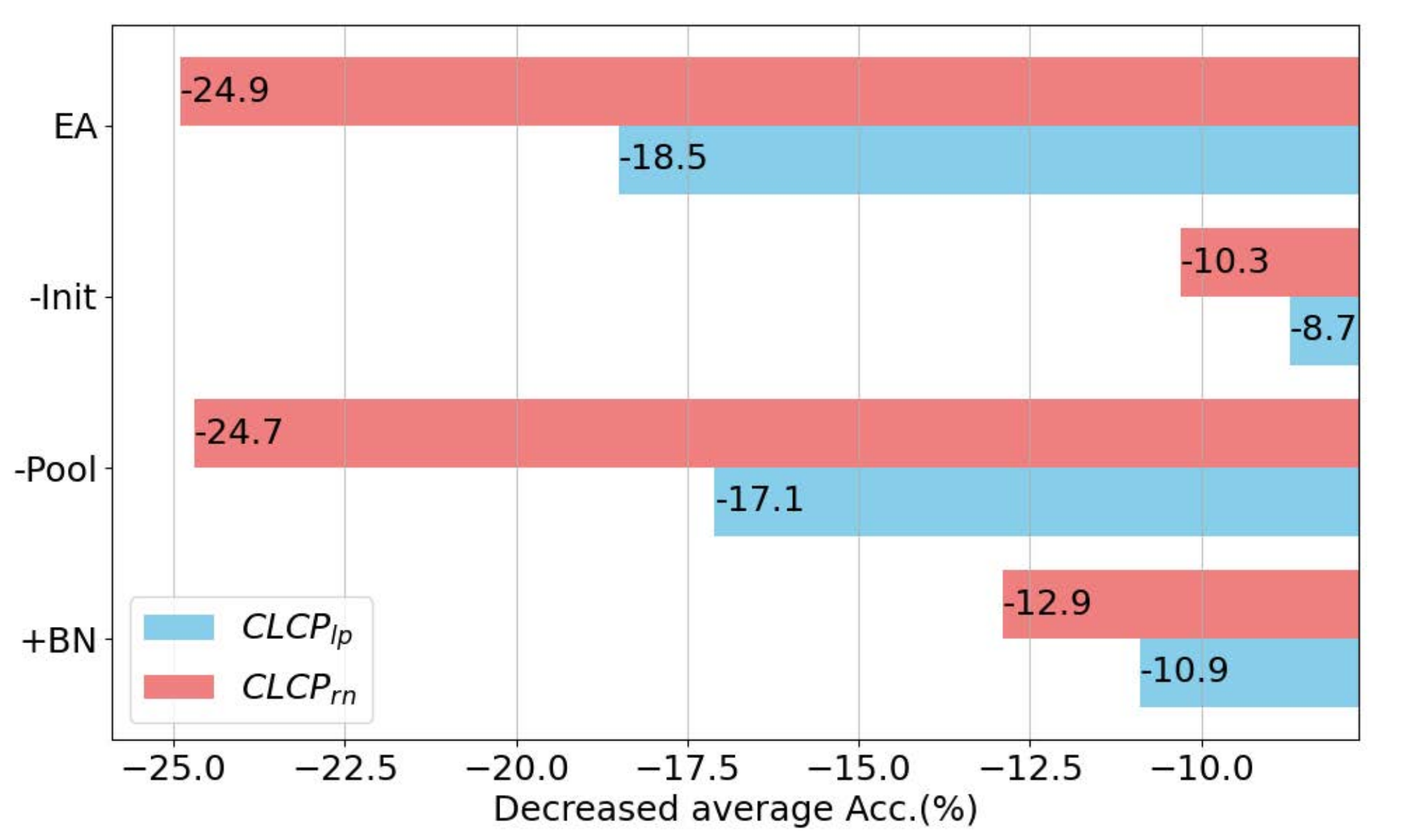}}
\caption{Removing and adding different components, changes in the effectiveness of CLCP$_{lp}$ and CLCP$_{rn}$.}
\label{RQ2}
\end{figure}

\textbf{Answer to RQ2:}
Fig. \ref{RQ2} shows the decrease in accuracy of models CLCP$_{lp}$ and CLCP$_{rn}$ compared to the original model when adding or removing different components.
EA represents the estimated Acc of the baseline random prediction.
The vertical axis represents adding (+) or deleting (-) components, and the horizontal axis represents decreased average ACC. of CLCP$_{lp}$ and CLCP$_{rn}$ with different sizes across all different training/testing dataset sizes. 
The model size and dataset size are the same as that of RQ1.
Legends of different colors represent different original models.

As shown in fig. \ref{RQ2}, removing the pooling layer did not alleviate underfitting as expected, but resulted in a significant decrease in accuracy.
The accuracy of \emph{CLCP$_{lp}$-Pool} is even lower than that of random prediction.
We observed that after $2$-$3$ epochs of training, the validation loss of \emph{CLCP$_{lp}$-Pool} and \emph{CLCP$_{rn}$-Pool} no longer decreased, which means that the training process of the models fell into local optima.
Therefore, after removing the pooling layer, the model experienced overfitting.
In addition, as expected, after removing He initialization, the performance of both \emph{CLCP$_{lp}$-Init} and \emph{CLCP$_{rn}$-Init} decreased compared to the original model.
After adding the BN layer, the performance of CLCP$_{lp}$ and CLCP$_{rn}$ also decreased.
This may because BN independently normalizes each batch of code feature maps, which may disrupt the relative differences between feature maps.

\begin{figure}[!htbp]
\centerline{
\includegraphics
[width=0.35\textwidth]
{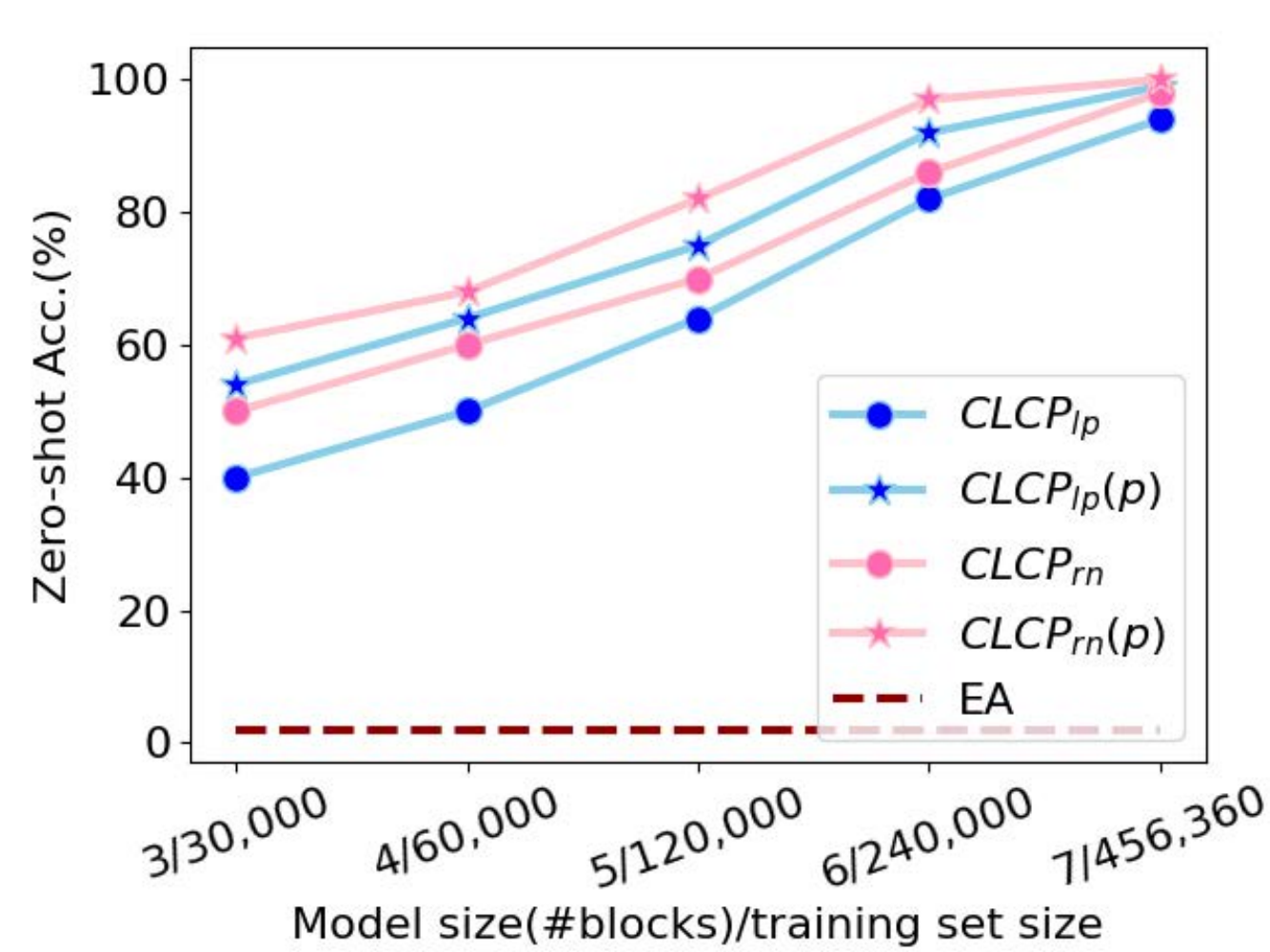}}
\caption{As the sizes of the training set and model increase, the performance of CLCP with and without prompt engineering on the testing sets of same size.}
\label{RQ2_sameTest}
\end{figure}

\begin{figure}[!htbp]
\centerline{
\includegraphics
[width=0.35\textwidth]
{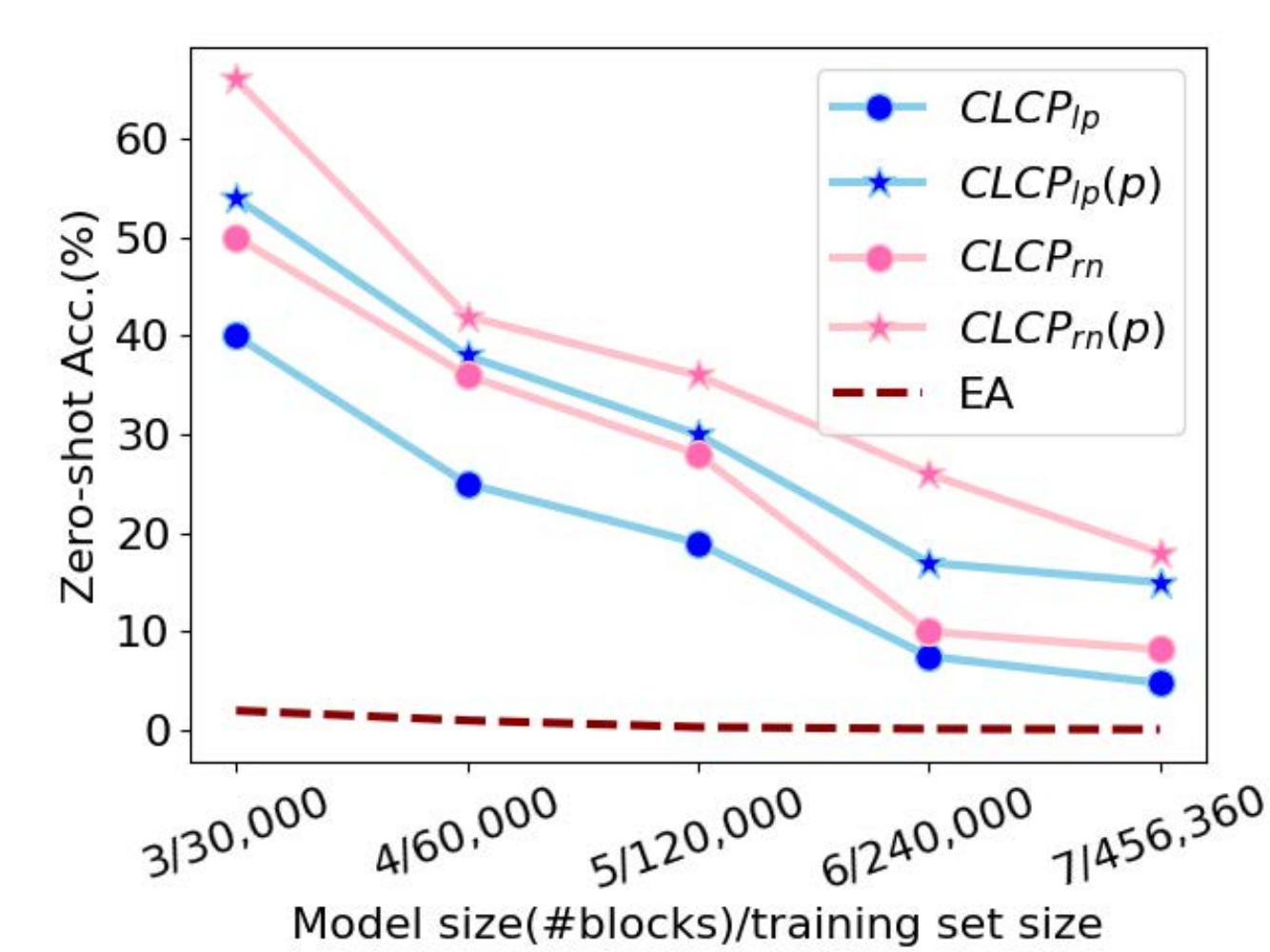}}
\caption{As the sizes of the training set and model increase, the performance of CLCP with and without prompt engineering on the testing sets of different size.}
\label{RQ2_diffTest}
\end{figure}

\textbf{Answer to RQ3:}
As the sizes of the training set and model increase, the performance of CLCP with and without prompt engineering on the testing sets of same and different size are show in Fig. \ref{RQ2_sameTest} and \ref{RQ2_diffTest}.
The overall performance of CLCP without prompt engineering (CLCP$_{lp}$ and CLCP$_{rn}$) is worse than that of CLCP with prompt engineering (CLCP$_{lp}$(p) and CLCP$_{rn}$(p)).
With an increase in data volume and models, the effectiveness of all models on the same testing dataset is steadily improving. In contrast, the effectiveness of all models decreases when the sizes of the testing dataset increase, regardless of whether prompt engineering is used. 
Therefore, using prompt engineering on the dataset can improve the performance of the model to some extent.

\section{Threat to Validity}
\textbf{Internal validity.}
Due to hardware and data limitations, we did not use as many datasets as OpenAI used to train CLIP models to train CLCP models. We simply followed CLIP's previous work of proposing text-to-image (which was only initially trained and tested on limited data) and proposed an initial text-to-code work. This results in the trained CLCP being not robust and cannot be directly used for downstream tasks of code understanding.
The main purpose of this paper is to provide a new perspective for researchers engaged in code understanding. In future work, we will collect more datasets and apply for more funding to evaluate our model on better hardware environments. And because we did not use pre-trained embeddings for code encoding, but instead proposed a new encoding paradigm and retrained it, we did not compare with other works that used ready-made embeddings.
\textbf{External validity.}
The source code of clip model is not open, and many training details have not been published.
Therefore, we replicated and improved its framework ourselves, which to some extent reduced the performance of proposed model.

\section*{Conclusion}
Microsoft Research and Google DeepMind have found many limitations in the GPTs' autoregressive paradigm, manifested in the model's lack of planning, working memory, backtracking, and reasoning skills.
Microsoft Research found that GPT relies on a local greedy process of generating the next word, without a global understanding of the task or output. We further summarized the limitations of the autoregressive paradigm on code understanding through empirical research. That is to say, GPTs cannot handle complex logic and generate new code that has not been seen before, and they rely too much n othe format of prompts to generate correct code. To address this issue, we propose a new code encoding paradigm inspired by the successful application of diffusion techniques in image generation, such as the Dalle2 and Sora models. We first discovered that the structure of code has both natural language characteristics and image/protein molecular chain characteristics, and encoded the code into a heterogeneous image paradigm with global information memory that mimics the structure of images and proteins. Then, we refer to Sora's CLIP upstream text-to-image encoder model and design a text-to-code encoder model that can be applied to various downstream code understanding tasks. This model learns the global understanding of code under the new paradigm, connects the encoding space of text and code, and encodes text input into the most similar code vector.
By conducting self-supervised comparative learning on 456360 text code pairs, the model achieved zero-shot prediction for new data.
This work is the foundation for using diffusion techniques to generate code in a new paradigm to avoid autoregressive limitations in the future.

% \section*{Data-Availability Statement}
% The artifacts are archived at: https://doi.org/10.5281/zenodo.13148594.

% \section*{Acknowledgment}

% The preferred spelling of the word ``acknowledgment'' in America is without 
% an ``e'' after the ``g''. Avoid the stilted expression ``one of us (R. B. 
% G.) thanks $\ldots$''. Instead, try ``R. B. G. thanks$\ldots$''. Put sponsor 
% acknowledgments in the unnumbered footnote on the first page.

\bibliographystyle{ieeetr}
\bibliography{Ref.bib}

\end{document}